\documentclass[11pt]{article}
\usepackage{eacl2017}
\usepackage{times}
\usepackage{url}
\usepackage{latexsym}
\usepackage{amsmath}
\usepackage{bm}
\usepackage{titlesec} 
\usepackage{graphicx}
\usepackage{booktabs} 

\eaclfinalcopy 

\title{Study of sampling methods in sentiment analysis of  imbalanced data}

\author{Zeeshan Ali Sayyed \\
  Department of Computer Science \\
  Indiana University \\
  Bloomington, IN 47405, USA \\
  {\tt zasayyed@indiana.edu}
  }

\date{\today}

\begin{document}
\maketitle
\begin{abstract}
  This work investigates the application of sampling methods for sentiment analysis on two different highly imbalanced datasets. One dataset contains online user reviews from the cooking platform Epicurious and the other contains comments given to the Planned Parenthood organization. In both these datasets, the classes of interest are rare. Word n-grams were used as features from these datasets. A feature selection technique based on information gain is first applied to reduce the number of features to a manageable space. A number of different sampling methods were then applied to mitigate the class imbalance problem which are then analyzed.
\end{abstract}

\section{Introduction}

Sentiment analysis deals with extracting opinions and/or emotions from from text, especially from the Internet. With the availability of large datasets from the growth of social media and other Internet technologies, it has become an important area of research \cite{INR-011} in the last decade.

Many real world datasets tend to have an imbalanced class distribution i.e. the proportion of samples from different classes is not uniform. High imbalance hampers the effectiveness of standard classification methods. This is because many standard algorithms assume a uniform class distribution and fail when presented with one which is heavily skewed. Therefore a lot of focus \cite{japkowicz2000learning} \cite{chawla2003workshop} has been put on developing techniques to combat the problem of imbalanced datasets. Machine learning on imbalanced datasets is called imbalanced learning. \cite{he2009learning} have compiled a summary of the techniques developed for imbalanced learning. Sampling is an important technique used by many researchers in imbalanced learning.

A survey of the literature reveals two important observations. Firstly, while a lot of work has been done on imbalanced learning in the binary classification setting, the problem of imbalanced learning in a multiclass classification setting has not been explicitly explored. Also, while a variety of techniques are used in the computational linguistics literature for tacking imbalanced learning in text classification and sentiment analysis, sampling has not been studied. One reason for this could be due to the the high dimensionality problem faced while dealing with text data. Since a lot of sampling techniques rely on distances between samples, this becomes computationally very expensive in a high dimensional space.

\section{Related Work}

Prediction of user ratings from online reviews was explored by \cite{yu:zhekova:ea:13}.
It was discovered that the classification is hampered by the class imbalance found in
the epicurious dataset described in sec \ref{sec:epicurious}. \cite{liu:guo:ea:14} and
\cite{liu:kuebler:ea:14} explored how this problem could be mitigated at the time of
feature selection. Finally, \cite{Kuebler_Liu_Sayyed_2018} proposed the use of feature
selection using multiclass information gain for feature selection in sentiment analysis
task.

Imbalanced learning is an active area of research. A brief survey of the the field is
given by \cite{he2013imbalanced}. There are four major categories of methods developed
for imbalanced learning:

\begin{itemize}
    \item Sampling methods
    \item Cost sensitive methods
    \item Kernel based methods
    \item Active learning methods
\end{itemize}

While sampling methods used in imbalanced learning have been described in section \ref{sec:sampling}, other methods will be covered in a future study.

\section{Problem Statement}
The aim of this study is to analyze the effect of using sampling techniques that have been developed for imbalanced learning on sentiment analysis of imbalanced data. In order to make it practical to apply sampling on high dimensional data, feature selection using multiclass information gain will be used to reduce the number of dimensions. The datasets which will be used in the study are multiclass. Most techniques have been studied on the binary class setting and hence this would test their robustness in a multiclass setting. Also, many techniques do not support multiclass settings inherently. So, this study also aims to modify them to enable them to work in a multiclass setting.

\section{Experimental Setup}

\subsection{Data Sets}

\begin{table}[t]
\begin{center}
\begin{tabular}{l|c|c|c}
Rating & No. of recipes(all) & In train & In test \\ \hline
 1 & 108 $(\approx 1.07\%)$ & 72 & 36\\
 2 & 787 $(\approx 7.8\%)$  &  522 & 265\\
3 &  5~648 & 3~763 & 1~885 \\
4 &  3~546 &  2~365 & 1~181 \\ \hline
all & 10~089 & 6~722 & 3~467\\ 
\hline
\end{tabular}
\end{center}
\caption{The distribution of ratings in the Epicurious data set.}\label{tab:epidist}
\end{table}

This study focuses on two datasets, viz. epicurious dataset and the planned parenthood dataset. Both these datasets have multiple classes and have high class imbalance amongst them. 

\subsubsection{Epicurious dataset} \label{sec:epicurious}
This dataset used by \cite{yu:zhekova:ea:13} consists of user reviews for about 10089 recipes which were crawled from the cooking website Epicurious\footnote{http://www.epicurious.com/}. Every review has an associated rating of 1 through 4 with it. The higher the rating, the more positive are the reviews associated with that particular recipe. These numbers were given by the users themselves which means there is a chance of human error. In fact there are a few reviews where the rating and the review do not go hand in hand and are considered as noise. The dataset was divided into a train set and a test set with a ration of 2:1 while maintaining the proportions of various classes. The distribution of classes in the Epicurious dataset is shown in Table \ref{tab:epidist}. It can be noted that classes $1$ and $2$ are the minority classes. 

\begin{table*}[t]
\begin{center}
\begin{tabular}{c|c|l}
Sentiment & No. of comments & Description \\ \hline
1 & 188 & extremely valuable \\
2 & 100 & useful \\
3 & 25 $(\approx 6.47\%)$ & neither useful nor useless \\
4 & 23 $(\approx 5.95\%)$ & do more harm than good \\
5 & 50 & are evil, they should be shut down \\ \hline
all & 100
\end{tabular}
\end{center}
\caption{The distribution and meaning of comments in the Planned Parenthood data set}
\label{tab:ppdist}
\end{table*}

\subsubsection{Planned Parenthood dataset}
This dataset contains comments from 386 subjects about the Planned Parenthood organization. The dataset was collected at the University of Miami as part of a larger questionnaire. The users were first asked to rank Planned Parenthood and then explain what they thought about it. The ratings were given on a scale of $1$ to $5$. The description of these sentiments regarding the policies and services offered by Planned Parenthood and their distribution is shown in Table \ref{tab:ppdist}. Given the size of the dataset is not very large, a stratified $10$-fold cross validation was performed on this dataset with approximately $40$ samples reserved as test set in each fold.

\subsection{Data Preprocessing}
Since both the data sets are not formal, they contain a lot of instances of colloquial expressions, emoticons, etc. or misspellings. Moreover, some of the comments contains Emojois and sometimes certain non-English words.

In this work, the same preprocessing steps as that of \cite{kuebler:liu:ea:16} were applied on both the datasets. They are:
\begin{enumerate}
    \item Replace all URLs by a special token ``URL".
    \item Replace all emoticons by a special token ``EMO".
    \item Replace all numbers - integers, time intervals, fractions - by a special token ``KNUMK".
    \item Replace all words and punctuations with repeating letters by their original form. 
    \item Properly segment sentences, run-on punctuation, and run-on words.
    \item Separate all contractions into the first word and the contracted part. For example  ``don't" is separated into ``do" and ``n't".
\end{enumerate}

Filtering stopwords is a common practice but we do not do it in this study because it is assumed that they contain valuable information regarding the sentiment of the comment. For instance, a more formal comment which has a lot of punctuations might be indicative of a negative sentiment as people tend to write more formally when they express a negative opinion.

\subsection{Feature Representation}
The choice of selecting features from text data is an important one as it can influence the overall accuracy of the system. But since the objective of this study is to study the effect of sampling, the standard of bag-of-words approach is used for feature generation. This methodology was used in \cite{liu:kuebler:ea:14} and \cite{kuebler:liu:ea:16}. Every comment is represented by all $n$-grams where $1 \leq n \leq 3$. To avoid noise and overfitting, in case of the Epicurious dataset, all $n$-grams with frequences less than or equal to $4$ are removed whereas in case of the Planned Parentgood dataset, this threshold is set at $2$, owing to the small size of the dataset.

After this step, there are about 386,000 features in the Epicurious dataset and about 2500 features in the Planned Parenthood dataset.

\subsection{Feature Selection}
\cite{liu:kuebler:ea:14} have shown that IG is the most robust feature selection method in a multi-class prediction with extreme skewing. A multiclass information gain strategy to rank features is used in this study for choosing the top features. Information gain is calculated for every $n$-gram feature in both the datasets according to Eq. \ref{eq:multig} \cite{yang1997comparative}:

\begin{equation}
\label{eq:multig}
\begin{split}
    IG(f) = &\sum_{i=1}^{m}P(c_i)logP(c_i) + \\
            &\sum_{ f \in \{ 0,1\}} P(f) \sum_{i=1}^{m}P(c_i|f)logP(c_i|f) 
\end{split}
\end{equation}

where $m$ is the number of classes in the dataset and $c_i$ denotes a particular class. $f$ is the feature under consideration. This formula just considers the presence or absence of any feature $f$ and hence $f \in \{0, 1\}$.

This concept was extended and $f$ was allowed to represent the number of times a feature occurs instead of just its presence or absence. This allows for more information to be extracted from the features. Using this, every feature was assigned a value of information gain and the features were sorted accordingly.

Based on the work in \cite{kuebler:liu:ea:16}, about $2500$ \footnote{\label{tomek}When the features are sorted according to IG values, a few features get the same IG values. Hence to avoid randomly cutting off at $2500$ and discarding a few features with the same IG values, the IG threshold is moved so that there is separation between the features which are chosen from the ones which aren't chosen.} features were chosen from about $300,000$ features. Similarly, in order to exclude irrelevant features, about $1500$ features were chosen from $2500$ in the planned parenthood dataset.

\subsection{Classification algorithm}
Different feature selection methods and sampling techniques can behave differently in combination with different classification methods. Moreover various combinations of their parameters can also lead to different performances. But since, the main focus of this study is evaluate the performance of various sampling methods and not model learning or parameter optimization, it uses support vector machines for classification as they are widely used in text classification and sentiment analysis across the literature. The implementation by \cite{joachims1998making} for SVM multilcass (v2.2) was used with default parameters. \cite{kuebler:liu:ea:16} used a different version of SVM multiclass for more accuracy, but we use this boost learning speed as this is considerably faster to train..

\section{Sampling} \label{sec:sampling}
Sampling is one of the methods used in practice to counter the problem of class imbalance. It does so by trying to achieve a relatively balanced distribution of classes by modifying the dataset. The goal is not to have an exactly balanced distribution, but merely a distribution which the traditional classifiers are better able to handle. This project uses the implementations of sampling techniques by \cite{lemaitre2016imbalanced}.

In every method, the resultant ratio after sampling is defined as the number of samples in minority class over those in majority class. Therefore, a completely balanced dataset will have a ratio of 1. For all the methods which allow us to explicitly control this ratio, $10$ different ratios viz. $0.1, 0.2 \ldots 1.0$ were tried.

Methods which handle multiclass settings inherently are used as is, whereas the ones which don't (all oversampling methods) we propose two custom setting to enable them to handle multiple classes, namely `one vs all' and `one vs neighbor'.

In the `1 vs all' setting, sampling is performed $|c|$ times where $|c|$ is the number of classes in the dataset. For every run, the class under consideration is considered as the first class and all other classes are together considered as the second class. Thus the multiclass dataset is transformed into a binary dataset. Sampling is performed to achieve the desired ratio.

In the '1 vs neighbor` setting, the inherent ordering of classes is exploited. For instance class $1$ has class $2$ as its neighbor whereas class $3$ has classes $2$ and $4$ as its neighbours. Like in the previous setting, sampling is performed $|c|$ times where $|c|$ is the number of classes. For every run, the current class is considered as the first set and the neighboring classes are considered as the second set. The assumption is that this setting will help exploit the classification boundaries between neighboring classes.

\subsection{Random oversampling and undersampling}

These two methods are the most basic forms of sampling and serve as a baseline in our study. In random undersampling the majority class samples are discarded at random to achieve a more balanced distribution. In contrast, in random oversampling minority class samples are copied and repeated randomly until a more balanced class distribution is reached. While these two methods appear to be functionaly equivalent, they introduce their own set of problematic consequences which can be harmful for the classifier. Undersampling might lead to potential loss of information thus preventing the classifier from learning important concepts. On the other hand, repetition of instances in oversampling can cause overfitting. 

\subsection{Informed undersampling}
Informed undersampling methods attempt to mitigate the potential loss of information problem present in random undersampling. They attempt to retain most of the useful information present in the majority classes by only removing the redundant, noisy and/or borderline samples. Removing redundant samples is safe because they do not contribute any useful information to the classifier. Removal of noisy samples help make the classification robust and finally removal of borderline samples helps avoid perturbations in the classification boundary of the classifier.

The different informed undersampling methods evaluated in this study are:

\subsubsection{Condensed Nearest Neighbors}  \label{sec:cnn}
The condensed nearest neighbors rule (CNN rule) was first proposed by \cite{CondensedNearestNeighborPaper} as an improvement over the nearest neighbour rule (NN rule) for classification tasks. It preserves the basic approach of NN rule but forgoes the need to store all the points in a dataset to make a classification. This is done by summarizing all the points in the dataset by a representative set called the minimal consistent set. A consistent set is a subset of the original dataset whose decision boundary is identical to that of the original dataset. A minimal consistent set is a set of minimum number of points, which when used as stored reference set, will correctly classify all the remaining points in the given sample set. The decision boundary of the consistent set might or might not be identical to that of the original set. This method tends to retain points along the decision boundary which are important for good classification and remove points which are away from the boundary.

This technique is used to reduce the number of samples in majority class by only choosing those samples which fall in its minimal consistent set. In this method, we cannot directly control the ratio of the under sampling i.e. decide the new number of samples in the majority class. 

\subsubsection{Edited Nearest Neighbors} \label{sec:enn}
The edited nearest neighbor scheme by \cite{EditedNearestNeighborPaper} attempts to reduce the number of points in the training set in order to remove noisy samples and produce a smooth decision boundary. The algorithm proposed in the original paper is simple, yet powerful and belongs to a class of algorithms which make use of this concept. The original paper modifies the $K$-nearest neighbors algorithm as follows:

Let $( X_i, y_i)$ where $i = 1, 2, \cdots, N$ denote the samples in the set.
\begin{enumerate}
    \item For each $i$,
    \begin{enumerate}
        \item Find the $K$-nearest neighbors to $X_i$ among $\{X_1, X_2, \cdots, X_{i-1}, X_{i+1}, \cdots, X_N \}$
        \item Find the class $y_i$ by majority vote.
    \end{enumerate}
    \item Find the set of points which are misclassified by the above approach and remove them from the training set.
\end{enumerate}

\subsubsection{Repeated Edited Nearest Neighbor}
This method is a modification of the edited nearest neighbors scheme describe in the previous section.\cite{rennPaper} proposed to repeat the method until there are no more misclassified samples.

\subsubsection{One Sided Selection} 
\cite{OneSidedSelectionPaper} introduced a modified version of the condensed neighbor rule for highly imbalanced datasets where the minority class is important. Since the CNN rule works on the entire dataset, without regards to the various classes, it might also remove certain samples from the minority class. To preserve them this methods performs a one sided selection and only remove the examples from the majority class. This method also removes only the majority class samples from Tomek links \footnote{Take two examples $\bm{x}$ and $\bm{y}$ such that both belong to different classes. Let $\delta(\bm{x},\bm{y})$ denote the distance between them. The pair $(\bm{x}, \bm{y})$ is called a Tomek link if no example $\bm{z}$ exists such that $\delta(\bm{x},\bm{z}) \leq \delta(\bm{x},\bm{y})$ or $\delta(\bm{y},\bm{z}) \leq \delta(\bm{y},\bm{x})$}.

\subsubsection{Neighborhood Cleaning Rule}
The bsic idea of neighborhood cleaning rule proposed by \cite{NeighborhoodCleaningRulePaper} is the same as one sided selection. Both the methods try to preserve the important minority class and avoid reducing it's samples. The contrast though, is that it emphasizes more on data cleaning than simply on data reduction. To achieve this, this methods uses instead Wilson's edited nearest neighbor technique (see section \ref{sec:enn}) instead of Hart's condensed nearest neighbor (see section \ref{sec:cnn}) to down sample the majority class.

\subsubsection{Near Miss}
Three different near miss methods are described by \cite{NearMissPaper}. The first method was used in this study. In this method, only those samples from the majority class are chosen, which are close to some of the minority samples. This is done by calculating the average distance of a sample from the majority class to three closest minority class samples. This distance is used to order the samples from the majority class and select a certain amount. Thus in this method, we can directly control the ratio of under sampling. We define the ratio as number of samples in minority class over number of samples in majority class. Hence a range of different ratios from $0.1$ to $1.0$ (no sampling) were tried in this experiment. It can be observed that this is similar to other nearest neighbor techniques in that, it too model decision boundaries, but more locally.

\subsection{Over Sampling}
In order to avoid the risk of overfitting and as a result inability to generalize to unseen results, these methods do not simply replicate samples from the minority class. They generate samples synthetically which belong to the minority class.

All these techniques do not inherently support multi-class settings. In order to incorporate multiple classes, every algorithm was run in two different settings viz. 1 vs all and 1 vs neighbor.

In the 1 vs all setting, sampling is performed $|c|$ times where $|c|$ is the number of classes in the dataset. For every run, the class under consideration is considered as the first class and all other classes are considered as the second class, thus converting the multi-class dataset into a binary class dataset. Sampling is performed to increase the 
\subsubsection{SMOTE} 
Synthetic minority oversampling technique (SMOTE) generates artificial samples based on feature space similarities between the existing samples of the minority class. This probably creates good synthetic samples if the features are more meaningful, but in our studies a feature is binary and just represents the presence or absence of a word or word phrase. The various variants of SMOTE that have been tested in this study are:

\paragraph{Regular}
The regular version of SMOTE was developed by \cite{chawla2002smote}. In this method, for ever sample $\bm{x_i}$ in the minority class $S_{min}$, $K$-nearest neighbours are found for a given integer $K$. To create a synthetic sample, one of these $K$ nearest neighbours is selected randomly, say $\bm{\hat{x}_i}$. The new sample is calculated as:
\begin{equation}
    \bm{x_{new}} = \bm{x_i} + (\bm{\hat{x}_i} - \bm{x_i}) \times \delta
\end{equation}
where $\delta \in [0,1]$ is a random number

Thus the resulting sample lies on the line segment joining $\bm{x_i}$ and $\bm{\hat{x}_i}$. In this way, SMOTE augments the original data with meaningful samples from the minority class and helps improve classification.

\paragraph{Borderline} 
This method differs from the regular SMOTE in the way that it only generates minority samples near the classification boundary. \cite{han2005borderline} have based this idea on the assumption that, borderline samples are the ones which contribute the most towards classification.

The way it calculates borderline samples is by counting the number of $K$-nearest neighbors of the minority class sample which belong to the majority class. Let this number be $k'$. If $k' = K$, then all the neighbors of this minority class sample belong to the majority class and hence this sample is considered as noise and is ignored from sampling. If $0 \leq k' \leq \frac{K}{2}$, then the minority sample is considered safe and not considered in sampling. Whereas if $\frac{K}{2} \leq k' \leq K$, then the sample is considered as borderline and oversampling is performed just like the way it would be performed in SMOTE. This methods is called Borderline-SMOTE1.

There is another variant of the technique just describes and it is called Borderline-SMOTE2. This methods not only generates synthetic samples from each sample which is considered borderline, as above, but also does that from it's nearest neighbour belonging to the majority class. And this extra sample is also considered belonging to the minority class.

\paragraph{SVM} 
This method proposed by \cite{nguyen2011borderline} has the same basic idea as that of Borderline SMOTE discussed above, with two key differences in it's implementation. Instead of depending on the classes of the neighboring samples to find out the borderline, it is approximated by the support vectors obtained after training a standard SVM classfier. Moreover, there is one more key difference in the way synthetic samples are created. Unlike regular SMOTE and Borderline SMOTE, synthetic minority samples are generated not only on the line segment in between the the minority sample and one of it's neighbor, but also on outside this line segment by extrapolation towards the majority class. 

\subsubsection{ADASYN} \cite{he2008adasyn}
SMOTE runs the risk of over generalization because it generates the same number of synthetic samples for ever minority class sample without regards to its neighboring samples. Adaptive synthetic sampling tries to mitigate this by using a weighted distribution for different minority classes, where more data is generated for minority samples which are harder to classify and less for those that are easy.This is done by using a density distribution $\hat{r}_i$ to automatically decide the number of synthetic samples to be generated for every minority class sample. For every $\bm{x_i}$ that belongs to the minority class, we calculate $r_i$ as follows:
\begin{equation}
    r_i = \frac{k'_i}{K}
\end{equation}
where $k'_i$ is the number of majority class samples in $K$-nearest neighbors of $\bm{x_i}$. We then calculate $\hat{r}_i$ by normalizing $r_i$ so that $\hat{r}_i$ is a density distribution.
\begin{equation}
    \hat{r}_i = \frac{r_i}{\sum_i r_i}
\end{equation}

\subsection{Sampling with data cleaning techniques}
The philosophy of these methods as describes by \cite{batista2004study} is that class imbalance does not systematically hinder the performance of learning systems. The problem is actually learning too few minority class samples in presence of other complicating factors like class overlapping. Although oversampling helps in balancing the class distribution, it doesn't remove certain others problems. For instance, the class clusters might not be well defined if some majority class samples invade the minority class samples or vice versa. Such cases might lead to overfitting. The following two techniques combat this problem by applying known oversampling techniques with data cleaning techniques.

\subsubsection{SMOTE + Tomek links}
After applying SMOTE, this technique removes all the samples present in Tomek links, which was defined in footnote \ref{tomek}.

\subsubsection{SMOTE + ENN}
The motivation of this technique \cite{batista2003balancing} is the same as that of SMOTE + Tomek links. The different lies in the fact that ENN (see section \ref{sec:enn}) removes a lot more samples compared to Tomek links and hence an in depth data cleaning takes place.


\section{Results}

\subsection{Evaluation}
It has been shown throughout the literature on imbalanced learning that accuracy and error rate are very ineffective in the assessment of classification algorithms because the fail to take into consideration the imbalance in the classes. Hence we rely of f-score and precision and recall of individual classes in our multi-class classification experiments to evaluate the performance of various settings.

Moreover, for the planned parenthood dataset, where $10$-fold cross validation was performed we use the methodology proposed by \cite{forman2010apples} to calculate the aggregated f-score, precision and recall values across the $10$ folds.

\begin{figure}
    \centering
    \includegraphics[scale=0.55]{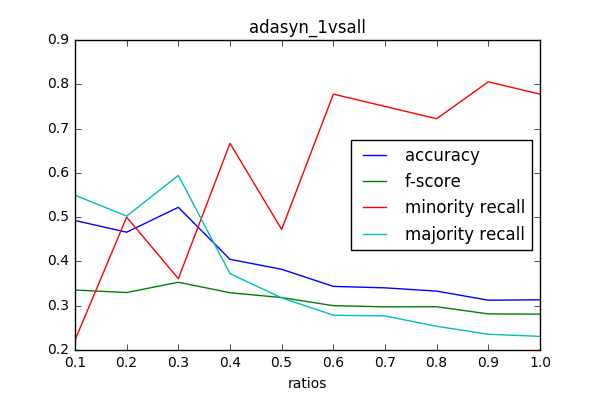}
    \caption{Variation of Accuracy, f-score and recalls of minority (1) and majority (3) class for adaptive synthetic sampling in a `1 vs all' setting on cooking dataset}
    \label{fig:ratio_variation}
\end{figure}

\subsection{Trends in varying ratios}
For every sampling method which supports explicit control of resulting ratio of minority class over majority class, $10$ different ratio values from $0.1$ to $1.0$ were tried. For instance, the adaptive synthetic sampling method in the `1 vs all' setting performs the best in the cooking dataset after one sided selection. Since multiple ratios were not tried in one sided selection due to lack of explicit control over the resultant ratio, we explore the adaptive synthetic sampling. 

Figure \ref{fig:ratio_variation} shows the variation in accuracy, f-score and recalls of majority/minority classes in adaptive synthetic sampling method in a `1 vs all' setting as the ratio of minority to majority class is varied from $0.1$ to $1.0$. The setting where the ratio is $0.3$ works best in terms of f-score and is the one that gets chosen as its representative. It should be noted that as the ratio approaches $1$ and the number of minority class samples approach that of majority class, the overall accuracy of the system goes down. It can be observed that the recall of minority class keeps on going up and that of majority class keeps on going down as the ratio increases. This is expected behavior as more synthetic samples are created for the minority classes and the classifier starts learning concepts from the minority classes. As the ratio goes up, many samples start getting classified as minority samples. After a point, many majority class samples also start getting classified as minority class as shown by the following recall of majority class.

For every sampling method, the ratio which gives the best performance in some measure like f-score is chosen as a representative of the ration.

\subsection{Cooking dataset}
Table \ref{tab:cookingresults} shows the accuracy, f-scores and precision \& recall values of all classes. The best ratio within a sampling technique is chosen on the basis of f-score values. The various techniques are ordered on the basis of their representative f-scores. It can be observed from the table that even though basic classification (without any sampling) gives higher accuracy, it falls behind a lot of other settings when f-scores are compared. In fact the precision and recalls values for minority class are $0$. This is because it does not learn concepts present in the minority samples. A lot of sampling methods can be seen to be having lower f-scores than the basic method. This is because they have very good precision/recall values for the minority classes but comparatively low values for the majority classes. This happens because sampling increases the importance of minority to such an extent that the classifier slightly overfits on them and hence does not perform well on the majority classes. One final thing to be noted is that with the exception of one sided selection technique, oversampling methods in general perform better than their undersampling counterparts for the cooking dataset.

\subsection{Planned Parenthood dataset}
Table \ref{tab:ppresults} shows the accuracy, f-scores and precision \& recall values of all classes. Similar to the cooking dataset, best ratio is chosen on the basic of f-score values and the various techniques are ordered on the basic f-score values. Unlike in the cooking dataset, the basic model without any sampling has very good f-score. We also observe that the various sampling techniques have good precision and recall values for the minority classes compared to the basic version. To investigate this further, we use recall of minority class (class 3) as the factor instead of f-score in deciding the representative ratio for a given sampling technique. The results are presented in table \ref{tab:ppresults2}. Also, even in this scenario, random undersampling performed the best. This shows that while sampling techniques were successful in helping the classifier achieve better precision and recall for the minority classes, the overall gain considering the precision and recall of the majority classes are not as good as the basic sampling. This is because as the classifier starts learning concepts from the minority samples, it overfits to it and starts losing information from the majority samples. Once again with the exception of one sided selection, the overall performance of oversampling methods is better than that of the undersampling ones. 

\begin{table*}
\resizebox{\textwidth}{!}{%

\begin{tabular}{lrrrrrrrrrr}
\toprule
                      name &  accuracy &   f-score &    prec\_1 &     rec\_1 &    prec\_2 &     rec\_2 &    prec\_3 &     rec\_3 &    prec\_4 &     rec\_4 \\
\midrule
         OneSidedSelection &  0.615385 &  0.361066 &  0.000000 &  0.000000 &  0.187050 &  0.098113 &  0.719702 &  0.614324 &  0.576623 &  0.751905 \\
             adasyn\_1vsall &  0.522127 &  0.352994 &  0.021036 &  0.361111 &  0.158672 &  0.162264 &  0.683761 &  0.594164 &  0.692857 &  0.492803 \\
           smote\_b2\_1vsall &  0.446094 &  0.349779 &  0.023209 &  0.638889 &  0.172727 &  0.286792 &  0.764835 &  0.369231 &  0.689084 &  0.598645 \\
          smote\_svm\_1vsall &  0.468666 &  0.349074 &  0.018539 &  0.472222 &  0.159314 &  0.245283 &  0.760365 &  0.486472 &  0.692584 &  0.490262 \\
           smote\_b1\_1vsall &  0.425601 &  0.348210 &  0.022090 &  0.722222 &  0.189815 &  0.309434 &  0.790433 &  0.368170 &  0.717045 &  0.534293 \\
        adasyn\_1vsneighbor &  0.463023 &  0.347878 &  0.018018 &  0.333333 &  0.143052 &  0.396226 &  0.759471 &  0.457294 &  0.697115 &  0.491109 \\
     smote\_svm\_1vsneighbor &  0.503416 &  0.346296 &  0.010526 &  0.194444 &  0.130120 &  0.203774 &  0.724162 &  0.561273 &  0.697337 &  0.487722 \\
      smote\_regular\_1vsall &  0.454707 &  0.346211 &  0.017192 &  0.333333 &  0.149471 &  0.426415 &  0.764599 &  0.444562 &  0.695226 &  0.480948 \\
 smote\_regular\_1vsneighbor &  0.454707 &  0.346211 &  0.017192 &  0.333333 &  0.149471 &  0.426415 &  0.764599 &  0.444562 &  0.695226 &  0.480948 \\
          RepeatedEditedNN &  0.546184 &  0.342618 &  0.000000 &  0.000000 &  0.085915 &  0.230189 &  0.707738 &  0.630769 &  0.688084 &  0.498730 \\
                     No sampling &  0.546184 &  0.342618 &  0.000000 &  0.000000 &  0.085915 &  0.230189 &  0.707738 &  0.630769 &  0.688084 &  0.498730 \\
      smoteenn\_1vsneighbor &  0.493020 &  0.341718 &  0.011665 &  0.305556 &  0.182927 &  0.113208 &  0.731396 &  0.547480 &  0.691402 &  0.497036 \\
    SMOTETomek\_1vsneighbor &  0.493020 &  0.341718 &  0.011665 &  0.305556 &  0.182927 &  0.113208 &  0.731396 &  0.547480 &  0.691402 &  0.497036 \\
      smote\_b2\_1vsneighbor &  0.549748 &  0.340880 &  0.006472 &  0.055556 &  0.084444 &  0.143396 &  0.699491 &  0.655703 &  0.683710 &  0.486876 \\
      smote\_b1\_1vsneighbor &  0.446985 &  0.340391 &  0.017734 &  0.500000 &  0.158151 &  0.245283 &  0.763815 &  0.454642 &  0.689866 &  0.478408 \\
                       RandomUnderSampling &  0.394713 &  0.312426 &  0.004587 &  0.055556 &  0.145060 &  0.781132 &  0.803977 &  0.300265 &  0.692500 &  0.469094 \\
                  NearMiss &  0.416988 &  0.311164 &  0.000000 &  0.000000 &  0.163296 &  0.822642 &  0.679710 &  0.248806 &  0.608659 &  0.607113 \\
                       NeighborhoodCleaningRule &  0.388773 &  0.305631 &  0.024256 &  0.611111 &  0.095471 &  0.294340 &  0.782511 &  0.370292 &  0.680426 &  0.432684 \\
           smoteenn\_1vsall &  0.341550 &  0.300990 &  0.017452 &  0.555556 &  0.157895 &  0.554717 &  0.811189 &  0.246154 &  0.722841 &  0.439458 \\
         SMOTETomek\_1vsall &  0.341847 &  0.300241 &  0.017327 &  0.583333 &  0.160151 &  0.479245 &  0.787826 &  0.240318 &  0.698856 &  0.465707 \\
   EditedNearestNeighbours &  0.345411 &  0.278493 &  0.021555 &  0.916667 &  0.076720 &  0.109434 &  0.796178 &  0.331565 &  0.707281 &  0.403048 \\
 CondensedNearestNeighbour &  0.502822 &  0.277129 &  0.000000 &  0.000000 &  0.057239 &  0.064151 &  0.578674 &  0.708223 &  0.713389 &  0.288738 \\
\bottomrule
\end{tabular}
}
\caption{Results of sampling methods on the cooking dataset. The ratio with best f-score was chosen as the representative of the sampling technique. The table is sorted according f-scores acRandomOversamplings sampling methods.}
\label{tab:cookingresults}
\end{table*}

\begin{table*}
\resizebox{\textwidth}{!}{%

\begin{tabular}{lrrrrrrrrrrrr}
\toprule
                      name &  accuracy &   f-score &    prec\_1 &     rec\_1 &    prec\_2 &  rec\_2 &    prec\_3 &  rec\_3 &    prec\_4 &     rec\_4 &    prec\_5 &  rec\_5 \\
\midrule
         OneSidedSelection &  0.585492 &  0.501319 &  0.813793 &  0.627660 &  0.540230 &   0.47 &  0.341463 &   0.56 &  0.179487 &  0.304348 &  0.540541 &   0.80 \\
                     No sampling &  0.575130 &  0.496747 &  0.804196 &  0.611702 &  0.541176 &   0.46 &  0.326531 &   0.64 &  0.179487 &  0.304348 &  0.542857 &   0.76 \\
          RepeatedEditedNN &  0.575130 &  0.496747 &  0.804196 &  0.611702 &  0.541176 &   0.46 &  0.326531 &   0.64 &  0.179487 &  0.304348 &  0.542857 &   0.76 \\
           smote\_b1\_1vsall &  0.551813 &  0.488682 &  0.804348 &  0.590426 &  0.512500 &   0.41 &  0.301887 &   0.64 &  0.207547 &  0.478261 &  0.548387 &   0.68 \\
             adasyn\_1vsall &  0.546632 &  0.486609 &  0.816794 &  0.569149 &  0.523810 &   0.44 &  0.313725 &   0.64 &  0.186441 &  0.478261 &  0.540984 &   0.66 \\
     smote\_svm\_1vsneighbor &  0.541451 &  0.485955 &  0.806202 &  0.553191 &  0.537500 &   0.43 &  0.290323 &   0.72 &  0.200000 &  0.478261 &  0.550000 &   0.66 \\
          smote\_svm\_1vsall &  0.551813 &  0.479508 &  0.800000 &  0.595745 &  0.512195 &   0.42 &  0.313725 &   0.64 &  0.170213 &  0.347826 &  0.530303 &   0.70 \\
        adasyn\_1vsneighbor &  0.541451 &  0.478024 &  0.809160 &  0.563830 &  0.517647 &   0.44 &  0.285714 &   0.56 &  0.192982 &  0.478261 &  0.531250 &   0.68 \\
      smote\_b2\_1vsneighbor &  0.546632 &  0.474694 &  0.800000 &  0.595745 &  0.500000 &   0.42 &  0.294118 &   0.60 &  0.183673 &  0.391304 &  0.532258 &   0.66 \\
      smote\_b1\_1vsneighbor &  0.544041 &  0.473058 &  0.805755 &  0.595745 &  0.519481 &   0.40 &  0.307692 &   0.64 &  0.166667 &  0.391304 &  0.515625 &   0.66 \\
 smote\_regular\_1vsneighbor &  0.536269 &  0.470452 &  0.800000 &  0.574468 &  0.500000 &   0.41 &  0.309091 &   0.68 &  0.180000 &  0.391304 &  0.500000 &   0.64 \\
      smote\_regular\_1vsall &  0.536269 &  0.470452 &  0.800000 &  0.574468 &  0.500000 &   0.41 &  0.309091 &   0.68 &  0.180000 &  0.391304 &  0.500000 &   0.64 \\
           smote\_b2\_1vsall &  0.544041 &  0.466968 &  0.808824 &  0.585106 &  0.505882 &   0.43 &  0.283019 &   0.60 &  0.152174 &  0.304348 &  0.530303 &   0.70 \\
                       RandomOversampling &  0.455959 &  0.426272 &  0.802198 &  0.388298 &  0.451613 &   0.42 &  0.253968 &   0.64 &  0.176471 &  0.521739 &  0.464789 &   0.66 \\
 CondensedNearestNeighbour &  0.440415 &  0.416437 &  0.750000 &  0.367021 &  0.443182 &   0.39 &  0.225806 &   0.56 &  0.157895 &  0.521739 &  0.529412 &   0.72 \\
      smoteenn\_1vsneighbor &  0.458549 &  0.413319 &  0.818182 &  0.478723 &  0.430380 &   0.34 &  0.257143 &   0.72 &  0.204819 &  0.739130 &  0.409091 &   0.36 \\
    SMOTETomek\_1vsneighbor &  0.458549 &  0.413319 &  0.818182 &  0.478723 &  0.430380 &   0.34 &  0.257143 &   0.72 &  0.204819 &  0.739130 &  0.409091 &   0.36 \\
                  NearMiss &  0.445596 &  0.409076 &  0.705882 &  0.319149 &  0.472727 &   0.52 &  0.229508 &   0.56 &  0.115385 &  0.260870 &  0.512821 &   0.80 \\
                       RandomUnderSampling &  0.443005 &  0.406864 &  0.779070 &  0.356383 &  0.466667 &   0.42 &  0.243243 &   0.72 &  0.083333 &  0.217391 &  0.513158 &   0.78 \\
           smoteenn\_1vsall &  0.432642 &  0.394789 &  0.793478 &  0.388298 &  0.493976 &   0.41 &  0.219178 &   0.64 &  0.136986 &  0.434783 &  0.415385 &   0.54 \\
         SMOTETomek\_1vsall &  0.419689 &  0.384439 &  0.793103 &  0.367021 &  0.471264 &   0.41 &  0.205479 &   0.60 &  0.135135 &  0.434783 &  0.415385 &   0.54 \\
                       NeighborhoodCleaningRule &  0.303109 &  0.255464 &  0.837838 &  0.164894 &  0.402778 &   0.58 &  0.026316 &   0.08 &  0.177570 &  0.826087 &  0.318182 &   0.14 \\
   EditedNearestNeighbours &  0.259067 &  0.204990 &  0.580000 &  0.154255 &  0.330827 &   0.44 &  0.034483 &   0.08 &  0.165468 &  1.000000 &  0.333333 &   0.04 \\
\bottomrule
\end{tabular}
}
\caption{Results of sampling methods on the planned parenthood dataset. The ratio with best f-score was chosen as the representative of the sampling technique. The table is sorted according f-scores across sampling methods.}
\label{tab:ppresults}
\end{table*}

\begin{table*}
\resizebox{\textwidth}{!}{%

\begin{tabular}{lrrrrrrrrrrrr}
\toprule
                      name &  accuracy &   f-score &    prec\_1 &     rec\_1 &    prec\_2 &  rec\_2 &    prec\_3 &  rec\_3 &    prec\_4 &     rec\_4 &    prec\_5 &  rec\_5 \\
\midrule
                       RandomUnderSampling &  0.417098 &  0.398481 &  0.752941 &  0.340426 &  0.426829 &   0.35 &  0.293333 &   0.88 &  0.142857 &  0.434783 &  0.405405 &   0.60 \\
        adasyn\_1vsneighbor &  0.409326 &  0.382228 &  0.773810 &  0.345745 &  0.438776 &   0.43 &  0.250000 &   0.84 &  0.168539 &  0.652174 &  0.451613 &   0.28 \\
                  NearMiss &  0.370466 &  0.360077 &  0.712500 &  0.303191 &  0.360000 &   0.27 &  0.283784 &   0.84 &  0.119048 &  0.434783 &  0.383562 &   0.56 \\
      smote\_b2\_1vsneighbor &  0.520725 &  0.470871 &  0.818182 &  0.526596 &  0.511905 &   0.43 &  0.303030 &   0.80 &  0.193548 &  0.521739 &  0.509434 &   0.54 \\
           smoteenn\_1vsall &  0.375648 &  0.342037 &  0.776471 &  0.351064 &  0.395062 &   0.32 &  0.204082 &   0.80 &  0.178947 &  0.739130 &  0.370370 &   0.20 \\
         SMOTETomek\_1vsall &  0.375648 &  0.342037 &  0.776471 &  0.351064 &  0.395062 &   0.32 &  0.204082 &   0.80 &  0.178947 &  0.739130 &  0.370370 &   0.20 \\
             adasyn\_1vsall &  0.518135 &  0.470731 &  0.798387 &  0.526596 &  0.493671 &   0.39 &  0.296875 &   0.76 &  0.203390 &  0.521739 &  0.516667 &   0.62 \\
      smote\_regular\_1vsall &  0.455959 &  0.430840 &  0.781250 &  0.398936 &  0.438202 &   0.39 &  0.279412 &   0.76 &  0.173913 &  0.521739 &  0.484375 &   0.62 \\
 smote\_regular\_1vsneighbor &  0.445596 &  0.405232 &  0.803922 &  0.436170 &  0.475000 &   0.38 &  0.253333 &   0.76 &  0.170732 &  0.608696 &  0.404255 &   0.38 \\
     smote\_svm\_1vsneighbor &  0.541451 &  0.485955 &  0.806202 &  0.553191 &  0.537500 &   0.43 &  0.290323 &   0.72 &  0.200000 &  0.478261 &  0.550000 &   0.66 \\
      smote\_b1\_1vsneighbor &  0.512953 &  0.460022 &  0.782258 &  0.515957 &  0.505747 &   0.44 &  0.272727 &   0.72 &  0.185185 &  0.434783 &  0.527273 &   0.58 \\
           smote\_b1\_1vsall &  0.461140 &  0.425900 &  0.794118 &  0.430851 &  0.411765 &   0.35 &  0.260870 &   0.72 &  0.163934 &  0.434783 &  0.492754 &   0.68 \\
      smoteenn\_1vsneighbor &  0.458549 &  0.413319 &  0.818182 &  0.478723 &  0.430380 &   0.34 &  0.257143 &   0.72 &  0.204819 &  0.739130 &  0.409091 &   0.36 \\
    SMOTETomek\_1vsneighbor &  0.458549 &  0.413319 &  0.818182 &  0.478723 &  0.430380 &   0.34 &  0.257143 &   0.72 &  0.204819 &  0.739130 &  0.409091 &   0.36 \\
          smote\_svm\_1vsall &  0.533679 &  0.473456 &  0.803030 &  0.563830 &  0.518987 &   0.41 &  0.293103 &   0.68 &  0.196429 &  0.478261 &  0.508197 &   0.62 \\
           smote\_b2\_1vsall &  0.455959 &  0.423426 &  0.787879 &  0.414894 &  0.435294 &   0.37 &  0.269841 &   0.68 &  0.161765 &  0.478261 &  0.464789 &   0.66 \\
                     No sampling &  0.575130 &  0.496747 &  0.804196 &  0.611702 &  0.541176 &   0.46 &  0.326531 &   0.64 &  0.179487 &  0.304348 &  0.542857 &   0.76 \\
          RepeatedEditedNN &  0.575130 &  0.496747 &  0.804196 &  0.611702 &  0.541176 &   0.46 &  0.326531 &   0.64 &  0.179487 &  0.304348 &  0.542857 &   0.76 \\
                       RandomOversampling &  0.455959 &  0.426272 &  0.802198 &  0.388298 &  0.451613 &   0.42 &  0.253968 &   0.64 &  0.176471 &  0.521739 &  0.464789 &   0.66 \\
         OneSidedSelection &  0.585492 &  0.501319 &  0.813793 &  0.627660 &  0.540230 &   0.47 &  0.341463 &   0.56 &  0.179487 &  0.304348 &  0.540541 &   0.80 \\
 CondensedNearestNeighbour &  0.440415 &  0.416437 &  0.750000 &  0.367021 &  0.443182 &   0.39 &  0.225806 &   0.56 &  0.157895 &  0.521739 &  0.529412 &   0.72 \\
                       NeighborhoodCleaningRule &  0.303109 &  0.255464 &  0.837838 &  0.164894 &  0.402778 &   0.58 &  0.026316 &   0.08 &  0.177570 &  0.826087 &  0.318182 &   0.14 \\
   EditedNearestNeighbours &  0.259067 &  0.204990 &  0.580000 &  0.154255 &  0.330827 &   0.44 &  0.034483 &   0.08 &  0.165468 &  1.000000 &  0.333333 &   0.04 \\
\bottomrule
\end{tabular}

}
\caption{Results of sampling methods on the planned parenthood dataset. The ratio with best rec\_3 was chosen as the representative of the sampling technique. The table is sorted according f-scores across sampling methods.}
\label{tab:ppresults2}
\end{table*}

\section{Conclusion}

In this paper we analyzed the role of sampling techniques in the combating class imbalance problem in two text data. In both the datasets the undersampling method of one sided selection performed the best in terms of f-score values. With this exception the various versions of smote and adaptive synthetic sampling which were extended to multiclass versions performed better than other undersampling methods. In general, sampling helped improve the precision and recall for minority classes in both the datasets. But in the small sized planned parenthood dataset, it did not have any significant contribution.

\bibliography{selection}
\bibliographystyle{eacl2017}

\end{document}